\documentclass[10pt,twocolumn,letterpaper]{article}

\usepackage{cvpr}              
\usepackage[normalem]{ulem}
\useunder{\uline}{\ul}{}
\usepackage{booktabs}
\usepackage{multirow}
\usepackage{graphicx}
\usepackage[accsupp]{axessibility} 
\usepackage[bottom]{footmisc}

\usepackage[dvipsnames]{xcolor}

\definecolor{cvprblue}{rgb}{0.21,0.49,0.74}
\usepackage[pagebackref,breaklinks,colorlinks,citecolor=cvprblue]{hyperref}

\def\paperID{****} 
\def\confName{CVPR}
\def\confYear{2024}

\newcommand{\mysmall}[1]{\scriptsize{\color{gray}{#1}}}

\title{SyncTalk: The Devil is in the Synchronization for Talking Head Synthesis}

\author{Ziqiao Peng$^1$\quad Wentao Hu$^2$\quad Yue Shi$^3$\quad Xiangyu Zhu$^4$ \quad Xiaomei Zhang$^4$ \quad Hao Zhao$^5$\\ Jun He$^1$\quad Hongyan Liu$^{5*}$\quad Zhaoxin Fan$^{1*}$ \\
$^1$Renmin University of China\quad $^2$Beijing University of Posts and Telecommunications \\ $^3$Psyche AI Inc.
\quad $^4$Chinese Academy of Sciences \quad $^5$Tsinghua University 
}

\begin{document}

\twocolumn[{
\maketitle
\begin{center}
    \captionsetup{type=figure}
    \vspace{-1.7em}
    \includegraphics[width=0.96\textwidth]{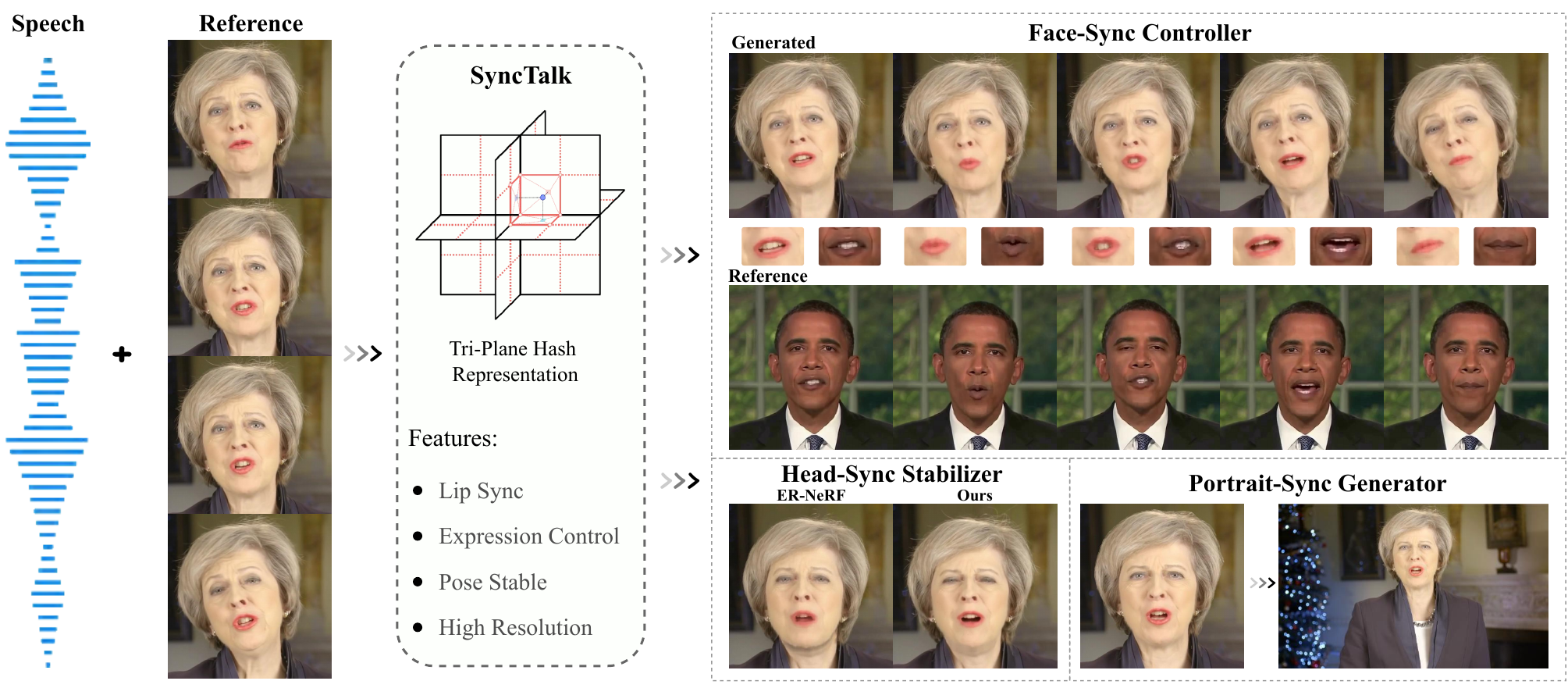}
    \vspace{-0.5em}
    \captionof{figure}{The proposed SyncTalk synthesizes synchronized talking head videos, employing tri-plane hash representations to maintain subject identity. It can generate synchronized lip movements, facial expressions, and stable head poses, and restores hair details to create high-resolution videos.}
    \label{fig:1}

\end{center}
}]

\renewcommand{\thefootnote}{\fnsymbol{footnote}} 
\footnotetext[1]{Corresponding authors.}

\begin{abstract}
Achieving high synchronization in the synthesis of realistic, speech-driven talking head videos presents a significant challenge. Traditional Generative Adversarial Networks (GAN) struggle to maintain consistent facial identity, while Neural Radiance Fields (NeRF) methods, although they can address this issue, often produce mismatched lip movements, inadequate facial expressions, and unstable head poses. A lifelike talking head requires synchronized coordination of subject identity, lip movements, facial expressions, and head poses. The absence of these synchronizations is a fundamental flaw, leading to unrealistic and artificial outcomes.
To address the critical issue of synchronization, identified as the ``devil'' in creating realistic talking heads, we introduce SyncTalk. This NeRF-based method effectively maintains subject identity, enhancing synchronization and realism in talking head synthesis. SyncTalk employs a Face-Sync Controller to align lip movements with speech and innovatively uses a 3D facial blendshape model to capture accurate facial expressions. Our Head-Sync Stabilizer optimizes head poses, achieving more natural head movements. The Portrait-Sync Generator restores hair details and blends the generated head with the torso for a seamless visual experience. Extensive experiments and user studies demonstrate that SyncTalk outperforms state-of-the-art methods in synchronization and realism. We recommend watching the supplementary video: \url{https://ziqiaopeng.github.io/synctalk}

\end{abstract}    
\section{Introduction}
\label{sec:intro}
The quest to generate dynamic and realistic speech-driven talking heads has intensified, driven by expanding applications in digital assistants~\cite{thies2020neural}, virtual reality~\cite{peng2023selftalk}, and film-making~\cite{kim2018deep}. The ultimate goal is to enhance the realism of synthetic videos to align with human perceptual expectations. However, a fundamental challenge that persists across existing methods is the need for synchronization. Traditional methods based on Generative Adversarial Networks (GAN)~\cite{zhang2023dinet,wang2023seeing,guan2023stylesync,zhong2023identity}, while adept at modeling the lip movements of speakers, often produce inconsistent identities across different frames, leading to issues such as different tooth sizes and fluctuating lip thickness. Similarly, emerging technologies based on Neural Radiance Fields (NeRF)~\cite{guo2021ad,yao2022dfa,shen2022learning,ye2023geneface,li2023efficient} excel in maintaining identity consistency and preserving facial details. However, they struggle with mismatches in lip movements, challenging facial expression control, and unstable head poses, thereby diminishing the overall realism of the video.

In this paper, we find that the ``devil'' is in the synchronization. Existing methods need more synchronization in four key areas: subject identity, lip movements, facial expressions, and head poses. Firstly, in GAN-based methods, maintaining the subject's identity in the video is challenging due to the instability of features in consecutive frames and the use of only a few frames as references for facial reconstruction~\cite{prajwal2020lip}. Secondly, lip movements are not synchronized with speech. In NeRF-based methods, audio features trained only on a 5-minute speech dataset struggle to generalize to different speech inputs~\cite{ye2023geneface}. Thirdly, there is a lack of facial expression control, with most methods only producing lip movements or controlling blinking, resulting in unnatural facial actions~\cite{gowda2023pixels}. Fourthly, the head pose is not synchronized. Previous methods relied on sparse landmarks to compute projection error, but jitter and inaccuracy in these landmarks lead to unstable head poses~\cite{zhang2021flow}, as shown in Fig.~\ref{fig:1}. These synchronization issues introduce artifacts and significantly reduce realism.

To address these synchronization challenges, we introduce SyncTalk, a NeRF-based method focused on highly synchronized, realistic, speech-driven talking head synthesis, employing tri-plane hash representations to maintain subject identity. Through the Face-Sync Controller and Head-Sync Stabilizer, SyncTalk significantly enhances the synchronization and visual quality of the synthesized videos. Visual quality is further improved by the Portrait-Sync Generator, which meticulously refines visual details. The entire rendering process can achieve 50 FPS and outputs high-resolution videos.

Within the Face-Sync Controller, we pre-train an audio-visual encoder on the 2D audio-visual dataset, resulting in a generalized representation that ensures synchronized lip movements across different speech samples. For controlling facial expressions, we employ a semantically enriched 3D facial blendshape model~\cite{peng2023emotalk}. This model is distinguished by its ability to control specific facial expression regions through 52 parameters. Regarding the Head-Sync Stabilizer, we use a head motion tracker~\cite{guo2021ad} to infer the head's rough rotation and translation parameters. Due to the instability of the rough parameters, inspired by Simultaneous Localization and Mapping (SLAM), we incorporate a head point tracker to track dense keypoints and integrate a bundle adjustment method to optimize the head pose, thereby achieving stable and continuous head motion.
To further enhance the visual fidelity of SyncTalk, we have designed a Portrait-Sync Generator. This module repairs artifacts in the NeRF modeling, particularly the intricate details of hair and background flaws, and outputs high-resolution video. 

In summary, the main contributions of our work are as follows:
\begin{itemize}
\item[$\bullet$] We present a Face-Sync Controller that utilizes an Audio-Visual Encoder in conjunction with a Facial Animation Capturer, ensuring accurate lip synchronization and dynamic facial expression rendering.

\item[$\bullet$] We introduce a Head-Sync Stabilizer that tracks head rotation and facial movement keypoints. Utilizing the bundle adjustment method, this Stabilizer guarantees smooth and synchronous head motion.

\item[$\bullet$] We design a Portrait-Sync Generator that improves visual fidelity by repairing artifacts in NeRF modeling and refining intricate details like hair and background in high-resolution videos.

\end{itemize}
\section{Related Work}
\label{sec:related}
\begin{figure*}
\vspace{-1em}
\begin{center}
   \includegraphics[width=1.\linewidth]{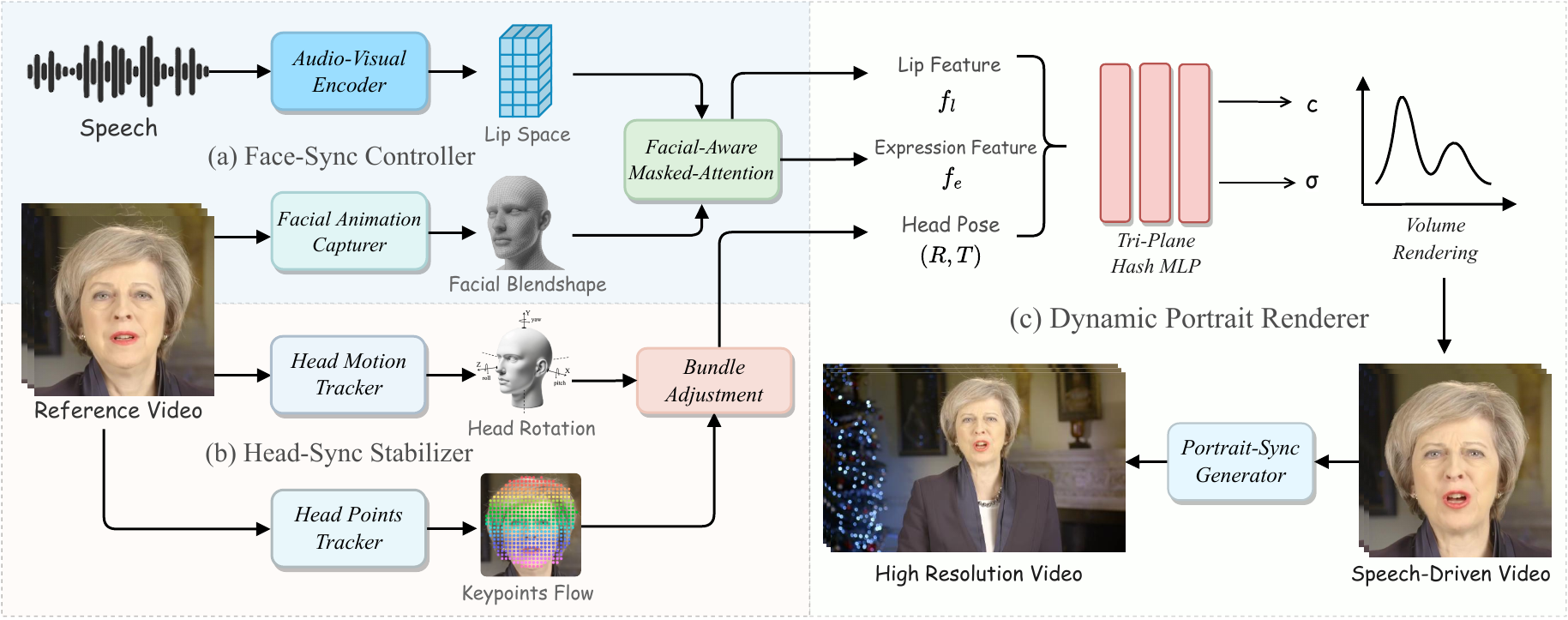}
\end{center}
\vspace{-1em}
   \caption{\textbf{Overview of SyncTalk.} Given a cropped reference video of a talking head and the corresponding speech, SyncTalk can extract the Lip Feature $f_l$, Expression Feature $f_e$, and Head Pose $(R, T)$ through two synchronization modules $(a)$ and $(b)$. The Tri-Plane Hash Representation then models the head, outputting a rough speech-driven video. The Portrait-Sync Generator further restores details such as hair and background, ultimately producing a high-resolution talking head video.}
\label{fig:2}
\vspace{-0.5em}
\end{figure*}
\subsection{GAN-based Talking Head Synthesis}

Recently, GAN-based talking head synthesis~\cite{chen2018lip,kr2019towards,chen2019hierarchical,zhou2019talking,das2020speech,vougioukas2020realistic,meshry2021learned,zhou2021pose,song2022everybody} has emerged as an essential research area in computer vision, but they struggle to maintain the identity of the subject in videos consistently.
A cluster of noteworthy techniques, including~\cite{prajwal2020lip,zhang2023dinet,wang2023seeing,sun2022masked,guan2023stylesync,zhong2023identity}, mainly focus on generating video streams for the lip region.
These methods create new visuals for talking head portraits by changing the lip region. For instance, Wav2Lip~\cite{prajwal2020lip} introduces a lip sync expert for supervising lip movements. However, due to the use of five frames from the reference frame to reconstruct the lip, it struggles to maintain the subject's identity. In contrast, methods like~\cite{chen2019hierarchical,zhou2020makelttalk,wang2021audio2head,lu2021live} perform full-face synthesis but struggle to ensure sync between facial expressions and head poses.

Apart from video stream techniques, efforts have also been made to enable a single image to ``speak'' using speech, as demonstrated in~\cite{zhang2021facial,ji2022eamm,zhang2023sadtalker}. For example, SadTalker~\cite{zhang2023sadtalker} can generate videos of a person speaking from a single image. However, such methods fail to generate natural head poses and facial expressions and struggle to maintain the subject's identity, affecting the sync effect and leading to an unrealistic visual perception.

Compared to these methods, SyncTalk uses NeRF to perform three-dimensional modeling of the face. Its capability to represent continuous 3D scenes in canonical spaces translates to exceptional performance in maintaining subject identity consistency and detail preservation.

\subsection{NeRF-based Talking Head Synthesis}
With the recent rise of NeRF, numerous fields have begun to utilize it to tackle related challenges~\cite{martin2021nerf,gao2022nerf}.
Previous work~\cite{guo2021ad,yao2022dfa,liu2022semantic,ye2023geneface} has integrated NeRF into the task of synthesizing talking heads and has used audio as the driving signal, but these methods are all based on the vanilla NeRF model. For instance, AD-NeRF~\cite{guo2021ad} requires approximately 10 seconds to render a single image. RAD-NeRF~\cite{tang2022real} aims for real-time video generation and employs a NeRF based on Instant-NGP~\cite{muller2022instant}. ER-NeRF~\cite{li2023efficient} innovatively introduces triple-plane hash encoders to trim the empty spatial regions, advocating for a compact and accelerated rendering approach. GeneFace~\cite{ye2023geneface} attempts to reduce NeRF artifacts by translating speech features into facial landmarks, but this often results in inaccurate lip movements. Attempts to create character avatars with NeRF-based methods, such as \cite{gafni2021dynamic,zheng2022avatar,zielonka2023instant,zheng2023pointavatar}, cannot be directly driven by speech.
These methods only use audio as a condition, without a clear concept of sync, and usually result in average lip movement. Additionally, previous methods lack control over facial expressions, being limited to controlling blinking only, and cannot model actions like raising eyebrows or frowning. Furthermore, these methods have a significant issue with unstable head poses, leading to the separation between the head and the torso.

In contrast, we use the Face-Sync Controller to model the relationship between audio and lip movements, thereby enhancing the synchronization of lip movements and expressions, and the Head-Sync Stabilizer to stabilize head poses. By addressing these synchronization ``devils'', our method improves visual quality.
\section{Method}
\label{sec:method}
\subsection{Overview}
In this section, we introduce the proposed SyncTalk, as shown in Fig.~\ref{fig:2}. SyncTalk mainly consists of 3 parts: a) lip movements and facial expressions controlled by the Face-Sync Controller, b) stable head pose provided by the Head-Sync Stabilizer, and c) high-synchronization facial frames rendered by the Dynamic Portrait Renderer. We will describe the content of these three parts in detail in the following subsections.

\subsection{Face-Sync Controller}
\noindent\textbf{Audio-Visual Encoder.} Existing methods based on NeRF utilize pre-trained models such as DeepSpeech~\cite{amodei2016deep}, Wav2Vec 2.0~\cite{baevski2020wav2vec}, or HuBERT~\cite{hsu2021hubert}. These are audio feature extraction methods designed for speech recognition tasks. Using an audio encoder designed for Automatic Speech Recognition (ASR) tasks does not truly reflect lip movements. This is because the pre-trained model is based on the distribution of features from audio to text, whereas we need the feature distribution from audio to lip movements.

Considering the above, we use an audio and visual synchronization audio encoder trained on the 2D audio-visual synchronization dataset LRS2~\cite{afouras2018deep}. This ensures that the audio features extracted by our method and lip movements have the same feature distribution. The specific implementation method is as follows: We use a pre-trained lip synchronization discriminator~\cite{chung2017out}. It can give confidence for the lip synchronization effect of the video. The lip synchronization discriminator takes as input a continuous face window $ F $ and the corresponding audio frame $ A $. They are judged as positive samples (with label $ y = 1 $) if they overlap entirely. Otherwise, they are judged as negative samples (with label $ y = 0 $). The discriminator calculates the cosine similarity between these sequences as:

\begin{equation}
\text{sim}(F, A) = \frac{F \cdot A}{\|F\|_2 \|A\|_2},
\end{equation}
and then uses binary cross-entropy loss:

\vspace{-1em}
\begin{equation}\label{eq:2}
L_{\text{sync}} = -\left( y \log(\text{sim}(F, A)) + (1-y) \log(1-\text{sim}(F, A)) \right),
\end{equation}
to minimize the distance for synchronized samples and maximize the distance for non-synchronized samples.

Under the supervision of the lip synchronization discriminator, we pre-train a highly synchronized audio-visual feature extractor related to lip movements. First, we use convolutional networks to obtain audio features $ \text{Conv}(A) $ and encode facial features $ \text{Conv}(F) $. These features are then concatenated. In the decoding phase, we use stacked convolutional layers to restore facial frames using the operation $\text{Dec}(\text{Conv}(A) \oplus \text{Conv}(F))$. The $L_1$ reconstruction loss during training is given by:   

\vspace{-1em}
\begin{equation}
L_{\text{recon}} = \|F - \text{Dec}(\text{Conv}(A) \oplus \text{Conv}(F))\|_1.
\end{equation}

Simultaneously, we sample synchronized and non-synchronized segments using lip movement discriminators, and employ a same sync loss as Eq.~\ref{eq:2}. By minimizing both losses, we train a facial generation network related to audio. After training, we use $ \text{Conv}(A) $ as the lip space extracted from the audio. Ultimately, we obtain our highly synchronized audio-visual encoder related to lip movements.

\begin{figure}
\begin{center}
   \includegraphics[width=1.\linewidth]{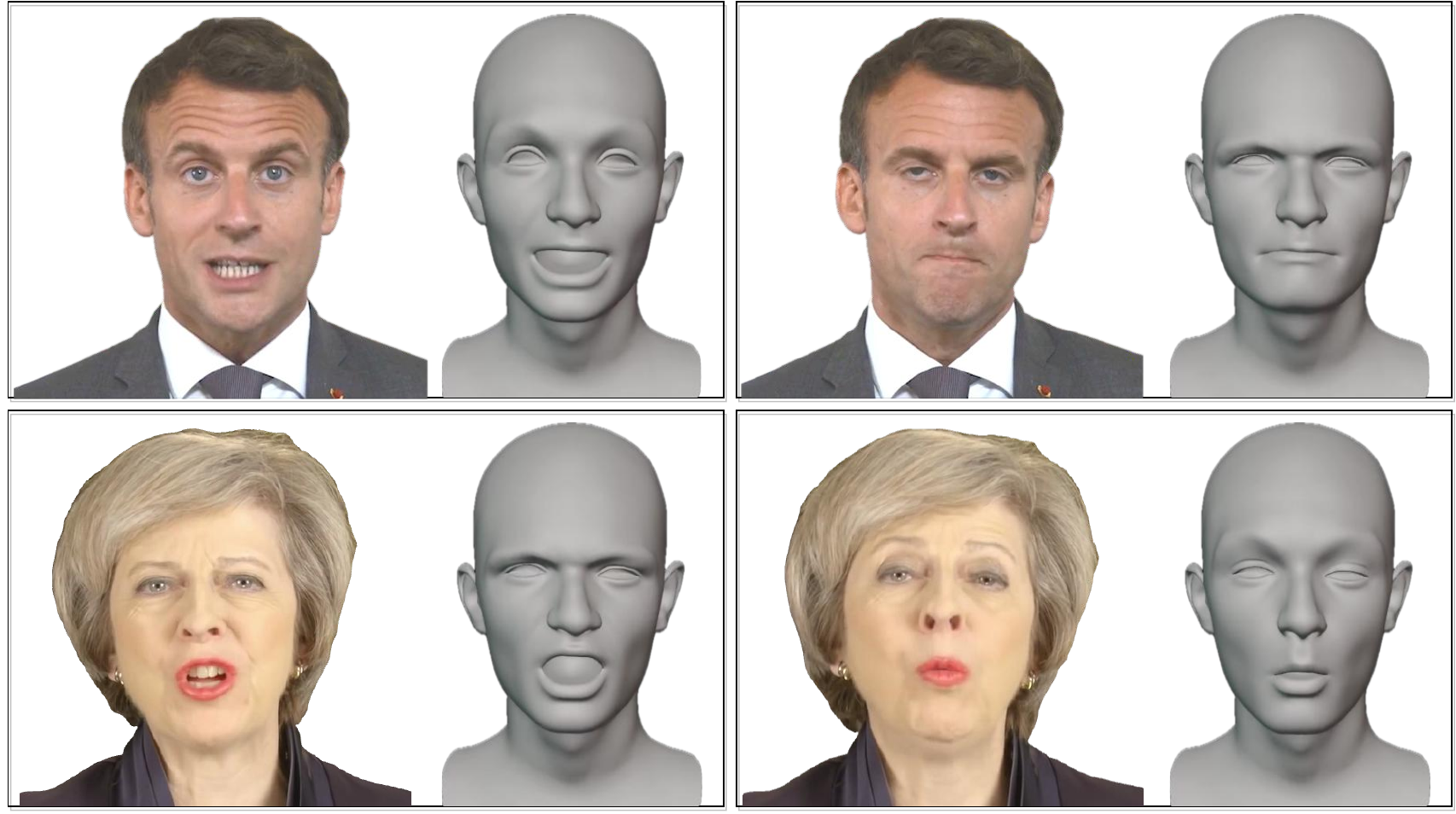}
\end{center}
\vspace{-1em}
   \caption{\textbf{Facial Animation Capturer.} We use 3D facial blendshape coefficients to control the expressions of characters.}
\label{fig:3}
\vspace{-1em}
\end{figure}

\noindent\textbf{Facial Animation Capturer.} It is observed that previous methods based on NeRF~\cite{guo2021ad,li2023efficient,ye2023geneface} could only change blinking and could not model facial expressions accurately. This leads to issues such as rigid facial expressions and incorrect facial details if the trained character has significant facial movements, like squinting, raising eyebrows, or frowning. Considering the need for more synchronized and realistic facial expressions, we add an expression synchronization control module. Specifically, we introduce a 3D facial prior using 52 semantically facial blendshape coefficients~\cite{peng2023emotalk} represented by $ B $ to model the face, as shown in Fig.~\ref{fig:3}. Because the 3D face model can retain the structure information of face motion, it can reflect the content of facial movements well without causing facial structural distortion. During the training, we first use a sophisticated facial blendshape capture module to capture facial expressions as $E(B)$, and select seven core facial expression control coefficients to control the eyebrow, forehead, and eye areas. They are highly correlated with expression and independent of lip movements. We can synchronize the speaker's expression during the inference process because the facial coefficients are semantically informed.

\noindent\textbf{Facial-Aware Masked-Attention.} To reduce the mutual interference between lip features and expression features during training, we introduce the Facial-Aware Disentangle Attention module. Building on the region attention vector $ V $~\cite{li2023efficient}, we add masks $ M_{\text{lip}} $ and $ M_{\text{exp}} $ to the attention areas for lips and expressions, respectively. Specifically, the new attention mechanisms are given by:

\vspace{-0.5em}
\begin{equation}
\begin{split}
V_{\text{lip}} &= V \odot M_{\text{lip}}, \\
V_{\text{exp}} &= V \odot M_{\text{exp}}.
\end{split}
\end{equation}

These formulations allow the attention mechanisms to focus solely on their respective parts, thereby reducing entanglement between them. Before the disentanglement, lip movements might induce blinking tendencies and affect hair volume. By introducing the mask module, the attention mechanism can focus on either expressions or lips without affecting other areas, thereby reducing the artifact caused by coupling. Finally, we obtain the disentangled lip feature ${f_l}=f_{\text{lip}} \odot V_{\text{lip}} $ and expression feature ${f_e} = f_{\text{exp}} \odot V_{\text{exp}}$.

\subsection{Head-Sync Stabilizer}
\noindent\textbf{Head Motion Tracker.} The head pose, denoted as $p$, refers to the rotation angle of a person's head in 3D space and is defined by a rotation $R$ and a translation $T$. An unstable head pose can lead to head jitter. To obtain a rough estimate of the head pose, initially, the best focal length is determined through $i$ iterations within a predetermined range. For each focal length candidate, $ f_i $, the system re-initializes the rotation and translation values. The objective is to minimize the error between the projected landmarks from the 3D Morphable Models (3DMM)~\cite{paysan20093d} and the actual landmarks in the video frame.
Formally, the optimal focal length $ f_{\text{opt}} $ is given by:
\begin{equation}
f_{\text{opt}} = \arg\min_{f_i} E_i(L_{2D}, L_{3D}(f_i, R_i, T_i)),
\end{equation}
where $ E_i $ represents the Mean Squared Error (MSE) between these landmarks, $ L_{3D}(f_i, R_i, T_i) $ represents the projected landmarks from the 3DMM for a given focal length $ f_i $, the corresponding rotation and translation parameters $ R_i $ and $ T_i $, $ L_{2D} $ are the actual landmarks from the video frame. 
Subsequently, leveraging the optimal focal length $ f_{\text{opt}} $, the system refines the rotation $ R $ and translation $ T $ parameters for all frames to better align the model's projected landmarks with the actual video landmarks. This refinement process can be mathematically represented as:
\begin{equation}
(R_{\text{opt}}, T_{\text{opt}}) = \arg\min_{R, T} E(L_{2D}, L_{3D}(f_{\text{opt}}, R, T)),
\end{equation}
where $ E $ denotes the MSE metric, between the 3D model's projected landmarks $ L_{3D} $ for the optimal focal length $ f_{\text{opt}} $, and the actual 2D landmarks $ L_{2D} $ in the video frame. The optimized rotation $ R_{\text{opt}} $ and translation $ T_{\text{opt}} $ are obtained by minimizing this error across all frames.

\begin{table*}[]
\setlength\tabcolsep{5pt}
\vspace{-1em}
\begin{center}
\resizebox{\linewidth}{!}{
\begin{tabular}{@{}c@{\hspace{6pt}}lcccc|cc|ccc@{}} 
\toprule
\multicolumn{2}{l}{Methods}                     & PSNR ↑           & LPIPS ↓         & MS-SSIM ↑       & FID ↓           & NIQE ↓           & BRISQUE ↓        & LMD ↓           & AUE ↓           & LSE-C ↑         \\ \midrule
\multirow{6}{*}{\rotatebox[origin=c]{90}{GAN}} & Wav2Lip \mysmall{(ACM MM 20 \cite{prajwal2020lip})} & 33.4385          & 0.0697          & 0.9781          & 16.0228         & 14.5367          & 44.2659          & 4.9630          & 2.9029          & \textbf{9.2387} \\
& \begin{tabular}[c]{@{}l@{}}VideoReTalking\\\mysmall{(SIGGRAPH Asia 22 \cite{cheng2022videoretalking})}\end{tabular}   & 31.7923          & 0.0488          & 0.9680          & 9.2063         & 14.2410          & 43.0465          & 5.8575          & 3.3308          & 7.9683          \\
& DINet \mysmall{(AAAI 23 \cite{zhang2023dinet})}     & 31.6475          & 0.0443          & 0.9640          & 9.4300          & 14.6850          & 40.3650          & 4.3725          & 3.6875          & 6.5653          \\
& TalkLip \mysmall{(CVPR 23 \cite{wang2023seeing})}   & 32.5154          & 0.0782          & 0.9697          & 18.4997         & 14.6385          & 46.6717          & 5.8605          & 2.9579          & 5.9472          \\
& IP-LAP \mysmall{(CVPR 23 \cite{zhong2023identity})}    & 35.1525          & 0.0443          & {\ul 0.9803}    & 8.2125          & 14.6400          & 42.0750      & 3.3350    & {\ul 2.8400}    & 4.9541                    \\ \midrule
\multirow{6}{*}{\rotatebox[origin=c]{90}{NeRF}} & AD-NeRF \mysmall{(ICCV 21 \cite{guo2021ad})}   & 26.7291          & 0.1536          & 0.9111          & 28.9862        & 14.9091          & 55.4667          & 2.9995          & 5.5481          & 4.4996          \\
& RAD-NeRF \mysmall{(arXiv 22 \cite{tang2022real})}    & 31.7754          & 0.0778          & 0.9452          & 8.6570          & {\ul 13.4433}    & 44.6892          & 2.9115          & 5.0958          & 5.5219          \\
& GeneFace \mysmall{(ICLR 23 \cite{ye2023geneface})}  & 24.8165          & 0.1178          & 0.8753          & 21.7084         & 13.3353          & 46.5061          & 4.2859          & 5.4527          & 5.1950          \\
& ER-NeRF \mysmall{(ICCV 23 \cite{li2023efficient})}   & 32.5216          & 0.0334          & 0.9501          & 5.2936          & 13.7048          & {\ul 34.7361}    & 2.8137          & 4.1873          & 5.7749          \\ \cmidrule(l){2-11} 
& SyncTalk (w/o Portrait)     & {\ul 35.3542}    & {\ul 0.0235}    & 0.9769          & {\ul 3.9247}    & \textbf{13.1333} & \textbf{33.2954} & {\ul 2.5714}    & \textbf{2.5796} & {\ul 8.1331}    \\
& SyncTalk (Portrait)       & \textbf{37.4017} & \textbf{0.0113} & \textbf{0.9841} & \textbf{2.7070} & 14.2165          & 37.3042          & \textbf{2.5043} & 3.2074          & 8.0263          \\ \bottomrule
\end{tabular}}
\end{center}
\vspace{-1em}
\caption{\textbf{The quantitative results of the head reconstruction.} ``Portrait'' refers to the use of the Portrait-Sync Generator. We achieve state-of-the-art performance on most metrics. We highlight \textbf{best} and \underline{second-best} results.}
\label{tab:1}
\vspace{-1em}
\end{table*}

\noindent\textbf{Head Points Tracker.} 
Considering methods based on NeRF and their requirements for inputting head rotation $ R $ and translation $ T $, previous methods utilize 3DMM-based techniques to extract head poses and generate an inaccurate result. To improve the precision of $R$ and $T$, We use an optical flow estimation model from~\cite{yao2022dfa} to track facial keypoints $ K $. Specifically, using a pre-trained optical flow estimation model, after obtaining the facial motion optical flow, we use the Laplacian filter to select the keypoints where the most significant flow changes are located and track the motion trajectories of these keypoints in the flow sequence. Through this module, our method ensures a more precise and consistent facial keypoint alignment across all frames, enhancing the accuracy of head pose parameters.

\noindent\textbf{Bundle Adjustment.} Given the keypoints and the rough head pose, we introduce a two-stage optimization framework from~\cite{guo2021ad} to enhance the accuracy of keypoints and head pose estimations. In the first stage, we randomly initialize the 3D coordinates of $j$ keypoints and optimize their positions to align with the tracked keypoints on the image plane. This process involve minimizing a loss function $ L_{\text{init}} $, which captures the discrepancy between projected keypoints $ P $ and the tracked keypoints $ K $, as given by:

\begin{equation} 
L_{\text{init}} = \sum_{j} \lVert P_j - K_j \rVert_2. 
\vspace{-0.5em}
\end{equation}

Subsequently, in the second stage, we embark on a more comprehensive optimization to refine the 3D keypoints and the associated head jointly pose parameters. Through the Adam Optimization~\cite{kingma2014adam}, the algorithm adjust the spatial coordinates, rotation angles $ R $, and translations $ T $ to minimize the alignment error $ L_{\text{sec}} $, expressed as:

\begin{equation} 
L_{\text{sec}} = \sum_{j} \lVert P_j(R, T) - K_j \rVert_2. 
\vspace{-0.5em}
\end{equation}

After these optimizations, the resultant head pose and translation parameters are observed to be smooth and stable.

\subsection{Dynamic Portrait Renderer}
\noindent\textbf{Tri-Plane Hash Representation.} Using a collection of multi-perspective images along with corresponding camera poses, NeRF \cite{mildenhall2021nerf} harnesses these resources to manifest a 3D static scene. This representation utilizes an implicit function $\mathcal{F}$, delineated as $\mathcal{F}:(\mathrm{x}, \mathrm{d}) \rightarrow (\mathrm{c}, \sigma),$
where the 3D spatial location is given by $\mathrm{x}=(x, y, z)$, and the direction of viewing is characterized by $\mathrm{d}=(\theta, \phi)$. The resultant values, $\mathrm{c}=(r, g, b)$ and $\sigma$, represent the radiance and density, respectively. The predicted pixel color, denoted by $\hat{C}(\mathrm{r})$, intertwined with ray $\mathrm{r}(t) = \mathrm{o}+t\mathrm{d}$ originating from the camera's core position $\mathrm{o}$, is derivable using:

\vspace{-0.2em}
\begin{equation}
\hat{C}(\mathrm{r}) = \int _{t_n}^{t_f}\sigma(\mathrm{r}(t)) \cdot \mathrm{c}(\mathrm{r}(t), \mathrm{d}) \cdot T(t)dt,
\vspace{-0.2em}
\end{equation}
where $t_n$ and $t_f$ are the near and far bounds, and $T(t)$ is the accumulated transmittance. Addressing the challenges of hash collisions and optimizing audio feature processing, we incorporate three uniquely oriented 2D hash grids \cite{li2023efficient}. A coordinate, given by $\mathrm{x}=(x, y, z)\in\mathbb{R}^{\mathrm{XYZ}}$, undergoes an encoding transformation for its projected values via three individual 2D-multiresolution hash encoders \cite{muller2022instant}:

\vspace{-0.5em}
\begin{equation}
\mathcal{H}^{\mathrm{AB}}: (a, b) \rightarrow \mathrm{f}_{ab}^{\mathrm{AB}},
\label{eq:planar_hash}
\end{equation}
where the output $\mathrm{f}_{ab}^{\mathrm{AB}} \in \mathbb{R}^{LD}$, with the number of levels $L$ and feature dimensions per entry $D$, signifies the planar geometric feature corresponding to the projected coordinate $(a, b)$ and $\mathcal{H}^{\mathrm{AB}}$ denotes the multiresolution hash encoder for plane $\mathbb{R}^{\mathrm{AB}}$.
By fusing the outcomes, the conclusive geometric feature $\mathrm{f}g \in \mathbb{R}^{3\times LD}$ is derived as:

\vspace{-1em}
\begin{equation}
\mathrm{f}_{\mathrm{x}} = \mathcal{H}^{\mathrm{XY}}(x,y) \oplus \mathcal{H}^{\mathrm{YZ}}(y,z) \oplus \mathcal{H}^{\mathrm{XZ}}(x,z),
\vspace{-0.5em}
\end{equation}
where the concatenation of features is symbolized by $\oplus$, resulting in a $3\times LD$-channel vector. Employing $\mathrm{f}_\mathrm{x}$, the perspective direction $\mathrm{d}$, the lip feature ${f_l}$, and the expression feature ${f_e}$, the tri-plane hash's implicit function is defined as:

\vspace{-0.5em}
\begin{equation}
\mathcal{F^H}: (\mathrm{x}, \mathrm{d}, {f_l}, {f_e}; \mathcal{H}^3) \rightarrow (\mathrm{c}, \sigma),
\end{equation}
where $\mathcal{H}^3$ amalgamates the triad of planar hash encoders, as illustrated in Eq.~\ref{eq:planar_hash}.

Our training employs a two-step coarse-to-fine strategy, initially using MSE loss to assess the difference between predicted $\hat{C}(\mathrm{r})$ and actual image colors $C(\mathrm{r})$. Recognizing MSE's limitations in detail capture, we advance to a refinement stage, incorporating LPIPS loss for enhanced detail, similar to ER-NeRF \cite{li2023efficient}. We extract random patches $ \mathcal{P} $ from the image, combining LPIPS (weighted by $ \lambda $) with MSE for improved detail representation, as shown:
\begin{equation}
\mathcal{L}_{\text{total}} = \sum_{\mathrm{r}} \lVert C(\mathrm{r}) - \hat{C}(\mathrm{r}) \rVert_2 + \lambda \times \mathcal{L}_{\text{LPIPS}}(\hat{\mathcal{P}}, \mathcal{P}).
\end{equation}

\noindent\textbf{Portrait-Sync Generator.} 
During the training process, to address NeRF's limitations in capturing fine details like hair strands and dynamic backgrounds, we introduce a Portrait-Sync Generator with two key sections. 
First, NeRF renders the face area ($ F_r $), creates $G (F_r) $through Gaussian blur, and then uses our synchronized head pose to be able to merge with the original image ($F_o $) to enhance hair detail fidelity.
Second, when the head and torso are combined, if the character in the source video speaks while the generated face is silent, a dark gap area might appear, as shown in Fig.~\ref{fig:5} $(b)$. We fill these areas with the average neck color ($ C_n $). This approach results in more realistic details and improved visual quality through the Portrait-Sync Generator.

\section{Experiments}
\label{sec:experiment}

\subsection{Experimental Settings}
\noindent\textbf{Dataset.} To ensure a fair comparison, we use the same well-edited video sequences from~\cite{guo2021ad,ye2023geneface,li2023efficient},  including English and French. The average length of these videos is approximately 8,843 frames, and each video is recorded at 25 FPS. Except for the video from AD-NeRF~\cite{guo2021ad}, which has a resolution of $450\times450$, all other videos have a resolution of $512\times512$, with the character-centered.

\noindent\textbf{Comparison Baselines.} We compare our method with five GAN-based methods, including Wav2Lip~\cite{prajwal2020lip}, VideoReTalking~\cite{cheng2022videoretalking}, DINet~\cite{zhang2023dinet}, TalkLip~\cite{wang2023seeing}, and IP-LAP~\cite{zhong2023identity}, and NeRF-based methods such as AD-NeRF~\cite{guo2021ad}, RAD-NeRF~\cite{tang2022real}, GeneFace~\cite{ye2023geneface}, and ER-NeRF~\cite{li2023efficient}.

\noindent\textbf{Implementation Details.} In the coarse stage, the portrait head is trained for 100,000 iterations and 25,000 in the fine stage, sampling $256^2$ rays per iteration using a 2D hash encoder ($L$=14, $F$=1). We employ the AdamW optimizer~\cite{loshchilov2017decoupled}, with learning rates of 0.01 for the hash encoder and 0.001 for other modules. Total training time is approximately 2 hours on an NVIDIA RTX 3090 GPU.

\begin{table}[]
\resizebox{\linewidth}{!}{
\begin{tabular}{@{}lcccc@{}}
\toprule
\multicolumn{1}{c}{\multirow{2.5}{*}{Methods}} & \multicolumn{2}{c}{Audio A}       & \multicolumn{2}{c}{Audio B}       \\ \cmidrule(l){2-5} 
\multicolumn{1}{c}{}                         & LSE-D ↓         & LSE-C ↑         & LSE-D ↓         & LSE-C ↑         \\ \midrule

DINet \mysmall{(AAAI 23 \cite{zhang2023dinet})}                                        & {\ul 8.5031}          & 5.6956          & {\ul 8.2038}          & 5.1134          \\
TalkLip \mysmall{(CVPR 23 \cite{wang2023seeing})}                                      & 8.7615          & {\ul 5.7449}          & 8.7019          & {\ul 5.5359}          \\
IP-LAP \mysmall{(CVPR 23 \cite{zhong2023identity})}                                       & 9.8037          & 3.8578          & 9.1102          & 4.389           \\
GeneFace \mysmall{(ICLR 23 \cite{ye2023geneface})}                                     & 9.5451          & 4.2933          & 9.6675          & 3.7342          \\
ER-NeRF \mysmall{(ICCV 23 \cite{li2023efficient})}                                      & 11.813          & 2.4076          & 10.7338         & 3.0242          \\
SyncTalk (Ours)                                     & \textbf{7.7211}    & \textbf{6.6659}    & \textbf{8.0248} & \textbf{6.2596} \\ \bottomrule
\end{tabular}}
\caption{\textbf{The quantitative results of the lip synchronization.} We use two different audio samples to drive the same subject, then highlight \textbf{best} and \underline{second-best} results.}
\label{tab:2}
\vspace{-1em}
\end{table}

\begin{figure*}
\vspace{-1.2em}
\begin{center}
   \includegraphics[width=1.\linewidth]{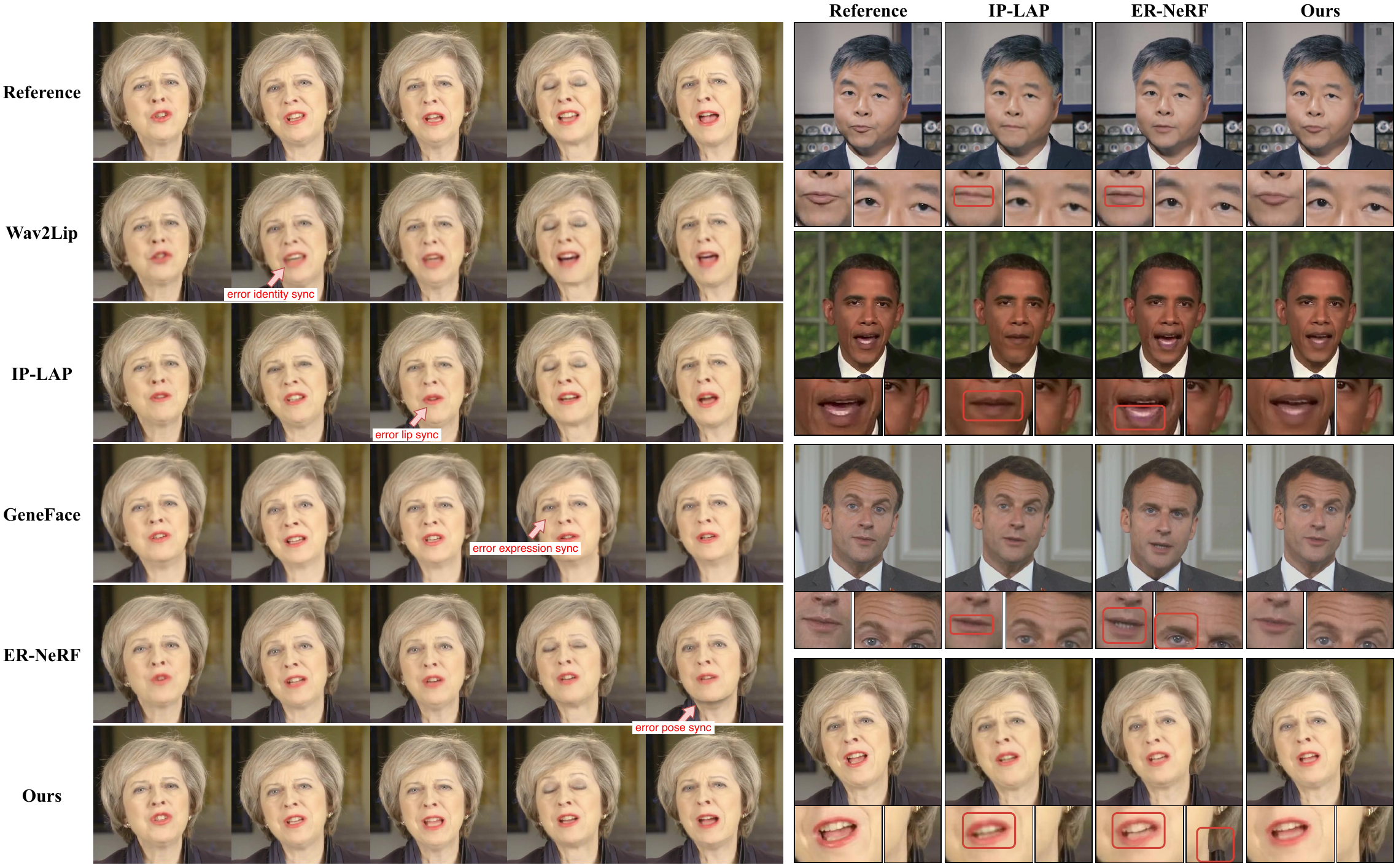}
\end{center}
\vspace{-1em}
   \caption{\textbf{Qualitative comparison of facial synthesis by different methods.} Our method has the best visual effect on lip movements and facial expressions without the problem of separation of head and torso. Please zoom in for better visualization.}
\label{fig:4}
\vspace{-1em}
\end{figure*}

\subsection{Quantitative Evaluation}
\noindent\textbf{Full Reference Quality Assessment.} In terms of image quality, we use full reference metrics such as Peak Signal-to-Noise Ratio (PSNR), Learned Perceptual Image Patch Similarity (LPIPS)~\cite{zhang2018unreasonable}, Multi-Scale Structure Similarity (MS-SSIM), and Frechet Inception Distance (FID)~\cite{heusel2017gans} as evaluation metrics. 

\noindent\textbf{No Reference Quality Assessment.} In high PSNR images, texture details may not align with human visual perception~\cite{zhang2019ranksrgan}. For more precise output definition and comparison, we use two No Reference methods: the Natural Image Quality Evaluator (NIQE)\cite{mittal2012making} and the Blind/Referenceless Image Spatial Quality Evaluator (BRISQUE)\cite{mittal2012no}.

\noindent\textbf{Synchronization Assessment.} For synchronization, we use landmark distance (LMD) to measure the synchronicity of facial movements, action units error (AUE)~\cite{baltruvsaitis2015cross} to assess the accuracy of facial movements, and introduce Lip Sync Error Confidence (LSE-C), consistent with Wav2Lip~\cite{prajwal2020lip}, to evaluate the synchronization between lip movements and audio.

\noindent\textbf{Evaluation Results.} The evaluation results of the head reconstruction are shown in Tab.~\ref{tab:1}. We compare recent methods based on GAN and NeRF. It can be observed that our image quality is superior to other methods in all aspects. In terms of synchronization, our results surpass most methods. We compare the two output modes of SyncTalk, one processed through the Portrait-Sync Generator and one without it. After processing through the Portrait-Sync Generator, hair details are restored, and the image quality is improved. Since we can maintain the subject's identity well, we surpass GAN-based methods in image quality. Thanks to the sync of lip, expression, and pose, we also outperform NeRF-based methods in image quality. Especially in terms of the LPIPS metric, our method has three times improvement compared to the previous state-of-the-art method ER-NeRF~\cite{li2023efficient}. We compare the latest SOTA method drivers using out-of-distribution (OOD) audio, and the results are shown in Tab.~\ref {tab:2}. We introduce Lip Synchronization Error Distance (LSE-D) and Confidence (LSE-C) for lip-audio sync evaluation, aligning with~\cite{prajwal2020lip}. Our method shows state-of-the-art lip synchronization, overcoming small-sample NeRF limitations by incorporating a pre-trained audio-visual encoder for lip modeling.

We also test the rendering speed. On an NVIDIA RTX 3090 GPU, and having preloaded the data onto the GPU, the head is outputted at 52 FPS with a resolution of $512 \times 512$. In Portrait mode, using the Portrait-Sync Generator, we achieve rendering speeds of 50 FPS with our CUDA acceleration. This far exceeds the video input speed of 25 FPS, allowing for real-time generation of video streams.

\subsection{Qualitative Evaluation}
\noindent\textbf{Evaluation Results.} 
To more intuitively evaluate image quality, we display a comparison between our method and other methods in Fig.~\ref{fig:4}. In this figure, it can be observed that SyncTalk demonstrates more precise and more accurate facial details. Compared to Wav2Lip~\cite{prajwal2020lip}, our method better preserves the subject's identity while offering higher fidelity and resolution. Against IP-LAP~\cite{zhong2023identity}, our method excels in lip shape synchronization, primarily due to the audio-visual consistency brought by the audio-visual encoder. Compared to GeneFace~\cite{ye2023geneface}, our method can accurately reproduce actions such as blinking and eyebrow-raising through expression sync. In contrast to ER-NeRF~\cite{li2023efficient}, our method avoids the separation between the head and body through the Pose-Sync Stabilizer and generates more accurate lip shapes. Our method achieves the best overall visual effect; we recommend watching the supplementary video for comparison.

\begin{table*}[]
\setlength\tabcolsep{2pt}
\begin{center}
\resizebox{\linewidth}{!}{
\begin{tabular}{@{}lcccccccccc@{}}
\toprule

                         & Wav2Lip~\cite{prajwal2020lip}  & DINet~\cite{zhang2023dinet} & TalkLip~\cite{wang2023seeing} & IP-LAP~\cite{zhong2023identity} & AD-NeRF~\cite{guo2021ad} & GeneFace~\cite{ye2023geneface} & ER-NeRF~\cite{li2023efficient} & SyncTalk      \\ \midrule
Lip-sync Accuracy        & {\ul 3.839}   & 3.696 & 2.893   & 3.161  & 2.696     & 2.982    & 3.189   & \textbf{4.304} \\
Exp-sync Accuracy & {\ul 3.536}   & 3.482 & 2.607   & 3.411  & 2.250   & 3.036    & 2.946   & \textbf{4.036} \\
Pose-sync Accuracy       & 3.571   & 3.571 & 2.875   & {\ul 3.696}  & 2.232     & 2.929    & 2.607   & \textbf{3.980} \\
Image Quality            & 2.500  & 2.696 & 2.054   & {\ul 3.571}  & 2.464    & 3.482    & 3.036   & \textbf{4.054} \\
Video Realness           & 2.929   & 2.429 & 2.429   & {\ul 3.161}  & 2.036    & 2.732    & 2.518   & \textbf{4.018} \\ \bottomrule
\end{tabular}
}
\end{center}
\vspace{-1em}
\caption{\textbf{User Study.} Rating is on a scale of 1-5; the higher the better. The term ``Exp-sync Accuracy'' is an abbreviation for ``Expression-sync Accuracy''. We highlight \textbf{best} and \underline{second-best} results.}
\label{tab:3}
\vspace{-1em}
\end{table*}

\begin{table}[]
\setlength\tabcolsep{3pt}
\renewcommand{\arraystretch}{1.1}

\begin{center}
\resizebox{\linewidth}{!}{
\begin{tabular}{@{}llll@{}}
\toprule
                                                                                          & PSNR ↑ & LPIPS ↓ & LMD ↓  \\ \midrule
Ours                                                                                      & \textbf{37.311} & \textbf{0.0121}  & \textbf{2.8032} \\ \midrule
\begin{tabular}[c]{@{}l@{}}replace Audio-Visual Encoder\\ with Hubert~\cite{hsu2021hubert}\end{tabular}     & 33.516 & 0.0276  & 3.3961 \\
\begin{tabular}[c]{@{}l@{}}replace Facial Animation Capture\\ with ER-NeRF~\cite{li2023efficient}'s\end{tabular} & 30.273 & 0.0415  & 3.0516 \\
\begin{tabular}[c]{@{}l@{}}w/o Facial-Aware\\ Masked-Attention\end{tabular}               & 36.536 & 0.0165  & 2.9139 \\
w/o Head-Sync Stabilizer                                                                  & 28.984 & 0.0634  & 3.5373 \\
w/o Portrait-Sync Generator                                                               & 32.239 & 0.0395  & 2.8154 \\ \bottomrule
\end{tabular}
}
\end{center}
\vspace{-1em}
\caption{\textbf{Ablation study for our components.} We show the PSNR, LPIPS, and LMD in different cases.}
\label{tab:4}
\vspace{-1em}
\end{table}

\noindent\textbf{User Study.} To provide a more comprehensive evaluation of the proposed model, we design an exhaustive user study questionnaire. We extract 24 video clips, each lasting more than 10 seconds, which include head poses, facial expressions, and lip movements. Each method is represented by three of these clips. We invite 35 participants to provide scores. The questionnaire is designed using the Mean Opinion Score (MOS) scoring protocol, asking participants to rate the generated videos from five perspectives: (1) Lip-sync Accuracy, (2) Expression-sync Accuracy, (3) Pose-sync Accuracy, (4) Image Quality, and (5) Video Realness. On average, participants take 19 minutes to complete the questionnaire, with a standardized Cronbach $\alpha$ coefficient of 0.96. The results of the User Study are displayed in Tab.~\ref{tab:3}. SyncTalk surpasses previous methods in all evaluations. Furthermore, we achieve the highest score in video authenticity, surpassing the second-place method, IP-LAP~\cite{zhong2023identity}, by a margin of 20$\%$. User studies indicate that our method can generate visually excellent quality as perceived by humans, achieving high realism.

\subsection{Ablation Study}
We conduct an ablation study to examine the contributions of different components in our model. We select three core metrics for evaluation: PSNR, LPIPS, and LMD. We select a subject named ``May'' for testing, and the results are presented in Tab.~\ref{tab:4}.

\begin{figure}
\begin{center}
   \includegraphics[width=1.\linewidth]{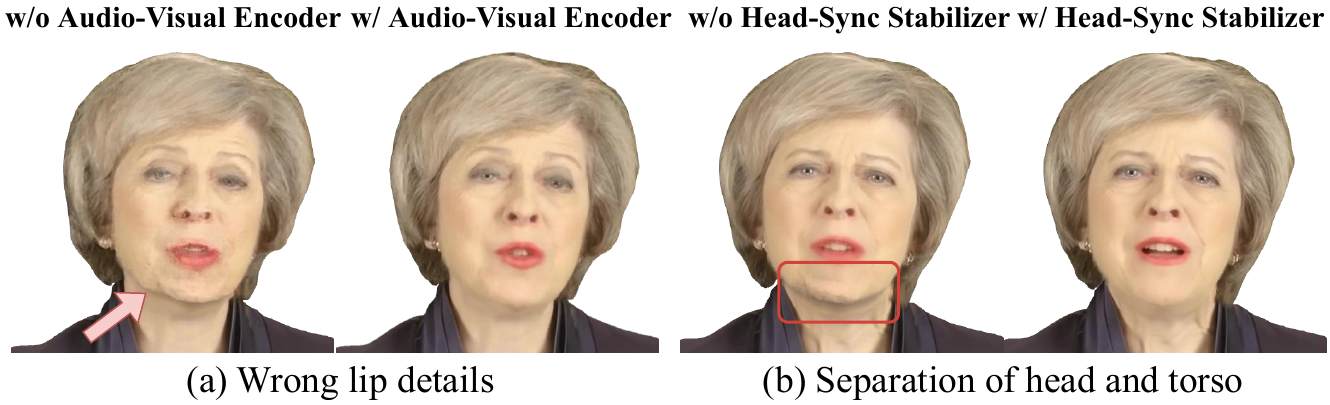}
\end{center}
\vspace{-1em}
   \caption{\textbf{Ablation Study} on Audio-Visual Encoder and Head-Sync Stabilizer. Removing them will lead to $(a)$ and $(b)$.}
\label{fig:5}
\vspace{-1em}
\end{figure}

The Audio-Visual Encoder provides the primary lip sync information. When this module is replaced, all three metrics become worse, with the LMD error increasing by 21.15$\%$ in particular, indicating decreased lip motion synchronization, as shown in Fig.~\ref{fig:5} $(a)$, and showing that our audio-visual encoders can extract accurate lip features. Replacing Facial Animation Capture with ER-NeRF~\cite{li2023efficient}'s blink module affects eyebrow movements and image quality.

The Facial-Aware Masked-Attention mainly alleviates motion entanglement between the lips and the rest of the face, slightly affecting the image quality after removal.
Without the Head-Sync Stabilizer, all metrics significantly declined, notably LPIPS, leading to head pose jitter and head and torso separation, as shown in Fig.~\ref{fig:5} $(b)$. The Portrait-Sync Generator restores details like hair. Removing this module impacts the restoration of details like hair, leading to noticeable segmentation boundaries.

\section{Conclusion}
\label{sec:conclusion}
In this paper, we introduce SyncTalk, a highly synchronized NeRF-based method for realistic speech-driven talking head synthesis. Our framework includes a Facial Sync Controller, Head Sync Stabilizer, and Portrait Sync Generator, which maintain subject identity and generate synchronized lip movements, facial expressions, and stable head poses. Through extensive evaluation, SyncTalk demonstrates superior performance in creating realistic and synchronized talking head videos compared to existing methods. We expect that SyncTalk will not only enhance various applications but also inspire further innovation in the field of talking head synthesis.

\section*{Acknowledgments}
This work was supported by National Natural Science Foundation of China (NSFC) under Grant Nos. 62172421 and 62072459, the Outstanding Innovative Talents Cultivation Funded Programs 2023 of Renmin University of China, and the Public Computing Cloud at Renmin University of China.

{
    \small
    \bibliographystyle{ieeenat_fullname}
    \bibliography{main}
}


\end{document}